\documentclass{article}

\usepackage{arxiv}

\usepackage[utf8]{inputenc} 
\usepackage[T1]{fontenc}    
\usepackage{hyperref}       
\usepackage{url}            
\usepackage{booktabs}       
\usepackage{amsfonts}       
\usepackage{nicefrac}       
\usepackage{microtype}      
\usepackage{lipsum}		
\usepackage{graphicx}
\usepackage{natbib}
\usepackage{doi}
\usepackage{subcaption}

\title{A Visual Analytics Approach to Building Logistic Regression Models and its Application to Health Records}

\author{ \href{https://orcid.org/0000-0002-2187-3949}{\includegraphics[scale=0.06]{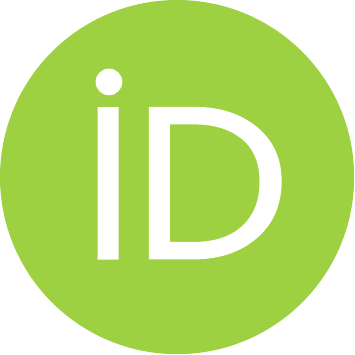}\hspace{1mm}Erasmo Artur}\thanks{Both authors contributed equally to this research.} \\
	Superintendence of Information Technology\\
	Federal University of Piauí\\
	Ininga SN, Teresina, Brazil \\
	\texttt{erasmo@ufpi.edu.br} \\
	\And
	\href{https://orcid.org/0000-0002-4799-8774}{\includegraphics[scale=0.06]{orcid.pdf}\hspace{1mm}Rosane Minghim} \\
	School of Computer Science and Information Technology\\
	University College Cork\\
	College Road, Cork, Ireland \\
	\texttt{r.minghim@cs.ucc.ie} \\
}

\renewcommand{\shorttitle}{\textit{arXiv} Template}

\hypersetup{
pdftitle={A template for the arxiv style},
pdfsubject={q-bio.NC, q-bio.QM},
pdfauthor={David S.~Hippocampus, Elias D.~Striatum},
pdfkeywords={First keyword, Second keyword, More},
}

\begin{document}
\maketitle

\begin{abstract}
Multidimensional data analysis has become increasingly important in many fields, mainly due to current vast data availability and the increasing demand to extract knowledge from it. In most applications the role of the final user is crucial to build proper machine learning models and to explain the patterns found in data. In this paper, we present an open unified approach for generating, evaluating, and applying regression models in high dimensional data sets within a user-guided process. The approach is based on exposing a broad correlation panorama for attributes, by which the user can select relevant attributes to build and evaluate prediction models for one or more contexts.  We name the approach UCReg (User-Centered Regression). We demonstrate effectiveness and efficiency of UCReg through the application of our framework to the analysis of Covid-19 and other synthetic and real health records data.
\end{abstract}

\keywords{regression analysis \and feature selection \and visual analytics \and attribute analysis}

\section{Introduction}
\label{Intro}

 Nowadays, large amounts of medical data (e.g., clinical information and treatment outcomes) have been stored in structured electronic health records (EHRs). These EHRs can represent a valuable source of information in an attempt to extract insights that can shape healthcare methods \cite{Goldstein:17}. A common way to use these data is by applying the predictive power of categorical attributes to evaluate probable outcomes supporting the decision-making process of the healthcare team and managers. In addition, medical studies are increasingly valuing the power of visual analysis methods \cite{Siggaard2020}. Medical research studies can be highly benefited by multidimensional visualization and multivariate prediction tools working in coordination. Regression analysis is a widely used method to investigate health records \cite{Goldstein:17, Dreiseitl:02}, where relevant attributes could potentially help building models for accurate predictions of outcomes. However, regression models generally do not deal well with many variables, mainly because of the multi-collinearity problem \cite{manly:16multi}, where strongly correlated attributes can destabilize the  model. Thus, a feature selection step plays an important role in the construction of a proper statistical model based on regression.

Various automatic methods for FS were proposed, as in \cite{battiti94using,duda12pattern,fleuret04fast,lin06conditional,liu95chi2,meyer06use,nie08trace,pedregosa11scikit,peng05feature,robnik03theoretical,yang99data,yu03feature}; however, excluding the analyst at this stage of the process may neglect the opportunity to employ her experience to the betterment of the resulting model. Automated algorithms may fail to recognize non-trivial relationships between attributes leading to low effectiveness in selection \cite{lu16recent}. Taking input from the analyst increases the chances of the selection being relevant thus producing a more accurate regression model.

It is also important to user guided methods that qualitative and quantitative evaluation of the models is performed to help improve model quality. Among many consolidated goodness-of-fit measures, a popular way to evaluate logistic regression (LR) efficacy is through the receiver-operating characteristic (ROC) curves \cite{metz78basic,hanley82meaning, hajian13receiver}, where the relationship between specificity and sensitivity is exposed among a range of cut-off values \cite{greiner00principles}. However, this is efficient to evaluate binary LR, where the goal is to determine the occurrence or not of an event. For multinomial LR, other solutions to evaluate the model's performance should be used, such as confusion matrices. To guide evaluation ov models we have developed a novel variant of the well established RadViz visualization technique \cite{Hoffman99RadViz} to provide an overview of the classification result by showing how the multinomial LR hits or misses the prediction.

\begin{figure}
	\includegraphics[width=13 cm]{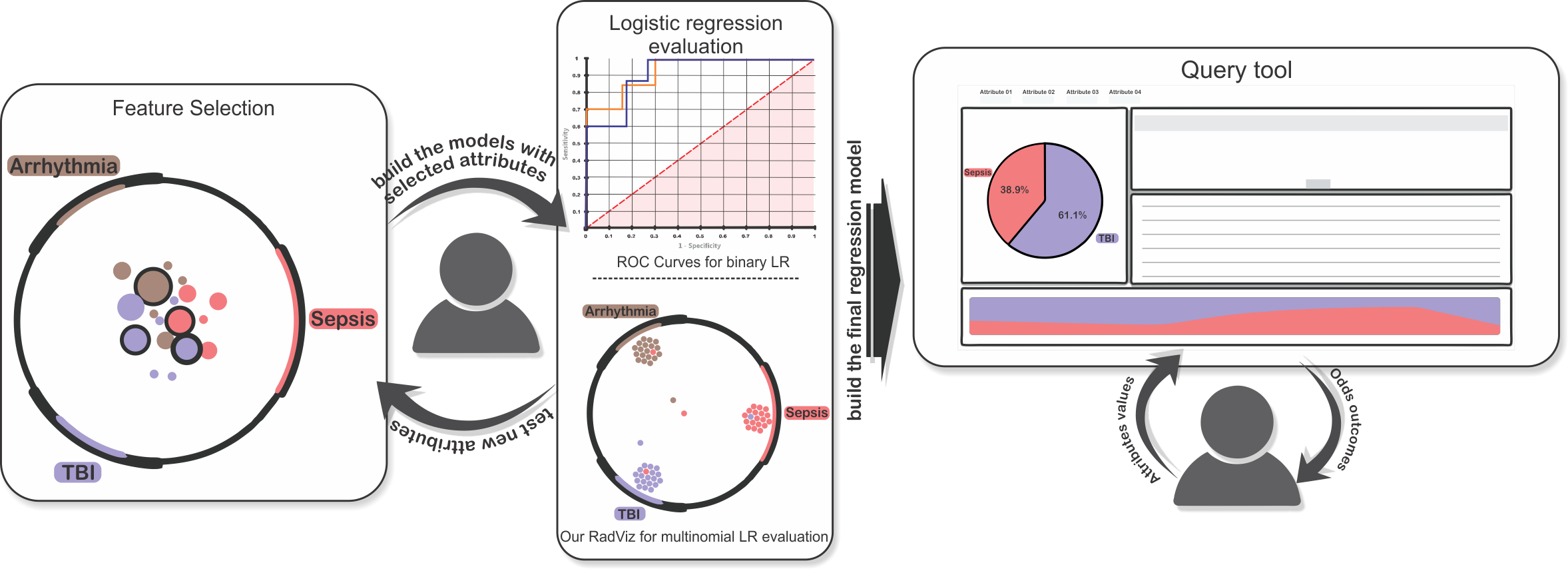}
	\centering
	\caption{Pipeline of the User-centered Regression (\textbf{UCReg}) approach. Initially, the analyst can interactively create and evaluate regression models. After gaining some knowledge, the analyst can create a tool for quick queries.}
	\label{fig:pipeline} 
\end{figure}

We describe in this paper a comprehensive approach for building and exploring LR models, and name UCReg(User-centered Regression). The approach, and the tool that implements it, include (see Figure \ref{fig:pipeline}): (i) feature analysis and selection, (ii) regression model construction, (iii) evaluating binary and multinomial regression, and (iv) constructing a panorama for queries over the model. The input of the approach is the data set containing the target attribute with the desired outcome labels. We employ a previously developed tool to provide inspection of attributes and their relationship to the target variable for the analyst to perform feature selection. From that the analyst chooses relevant variables to build a regression equation automatically. Evaluation is performed by a modified RadViz approach. Optionally, the analyst can create a query tool based on the generated regression models.

We demonstrate UCReg using medical records. We employ the trauma case to accompany the explanations due to its applicability in every step of the process. Then we show the results when realizing the processes on Covid-19 data at country and patient levels.

\section{Related Work}

Our User-Centered Regression approach and tool covers the whole pipeline of creating and evaluating a regression model guided by the model builder. From attribute correlation, to model building to evaluation, the method proposes and implements visual analysis tools to dive deep into data prediction steps. In this Section, we present an overview of previous wok on of these subjects' backgrounds and how we relate to them.

A key task in our approach is the interactive FS. Visual strategies are often employed to promote interactivity in this task, and the visualization community has devoted effort to creating tools to assist this process. An example of such work is the INFUSE (INteractive FeatUre SElection) present by Krause et al. \cite{krause14infuse}, a visual approach that provides an overview of the results of automatic FS algorithms. Another work present by Bernard et al. \cite{bernard2014visual} focus on the search for hidden correlations. The tool creates a relationship prospect between attributes and, more importantly, between their bins. Similarly, May et al. \cite{may11guiding} propose the SmartStripes, which supports the investigation of interdependencies between different attribute and entity subsets. They desire to find attributes with maximum correlation with instance partitions of the data set. Turkay et al. \cite{turkay11brushing} describe a linked dual view where item and dimension views are associated with an interactive model based on brushing and focus+context. Yuan et al. \cite{yuan13dimension} also present a dual view model with a simultaneous visual exploration of attributes and items spaces. The user can select the data by restricting its range as well as pruning dimensions, with the option of reducing its scope or representing the data by a scatter plot matrix or a tree. Recently, Rauber et al. \cite{rauber17projections} present a tool to support the classification systems design. Users can select attributes, project items according to the current selection, check the most relevant attributes for each label (or group), and investigate the relationship between attributes through a multidimensional projection. In this work, we adopt the Attribute-RadViz \cite{artur19radviz}, which builds a relationship panorama between attributes and data labels and presents them by applying the RadViz mapping. Our motivation in choosing the Attribute-RadViz lies in the generation of a meaningful overview of the relationships of attributes and their categories. In our approach, the user must be aware of the relevance of attributes concerning the data labels.

Once we provide an overview of attributes and interactive FS means, the approach then supports the exploration of regression models. Similarly,  other tools have been proposed aiming at the visual exploration of regression models. Piringer et al. \cite{piringer10hyp} describe the HyperMoVal, a framework for the task of validating regression models mainly by evaluating the combined visualizations of known and predicted data. M\"{u}hlbacher and Piringer \cite{muhlbacher13pbf} present a framework that initially ranks relevant attributes related to a target attribute (thought a chosen metric) and visually exposes the conditional distribution of the dependent attribute relative to the input attributes. Klemm et al. \cite{klemm16regheatmap} present a technique that visually displays combinations of regression models by varying independent attributes combinations related to a chosen target one. The approach also allows adjustment on regression formulas and metrics selection for analyzing the model's performance. These approaches generate (sometimes exhaustively) multiple regression models and visually expose their results. In our approach, a correlation analysis is responsible for presenting potentially good predictors (independent attributes) to the user, and then the models can be created and evaluated. Our focus is the support of model creation and exploration, and then the building of a query mechanism to take advantage of the generated models learning for later predictions of interest.

Although our approach has applicability in other areas, we have demonstrated our usefulness focusing on visual EHRs analysis. Several systems have also been proposed for the interactive visualization of EHRs. Rind et al. \cite{rind13eghvis} published reviews on a paper that cover most of them, while Gotz et al. \cite{gotz16ddhealth} focus on the challenges to be overcome by theses visual approaches. Patient visual cohort analysis has been the focus of approaches like the one present by Rogers et al. \cite{rogers19composer}, an Composer system that compares changes in the physical condition of patients after multiple spinal procedures. Bernard et al. \cite{bernard14visual} present an interactive visual tool for a cohort analysis of prostate cancer patients. Malik et al. \cite{malik15coco} propose the CoCo tool, which focuses on the comparison between cohorts of temporal event sequences. While these results have made significant contributions to the visualization of cohort patient records, the support for prediction and decision-making assistance remain open. Our approach allows users to visualize attributes correlations and explore regression models to gain insight and also investigate outcomes that may be predicted in the data set.

Recently proposed approaches combine attribute analysis with statistical models for prediction, commonly LR. Zhang et al. \cite{zhang16visual} present a method for creating LR models aided by visualization; initially, a FS based on a univariate analysis view is performed for the later model building and evaluation. The approach provides a attribute group view, which displays relevant attribute's information as distribution, correlation, and factor analysis, permitting users to deal with the multicollinearity problem. More similar to our approach, Dingen et al. \cite{dingen19reg} introduce the RegressionExplorer, a tool that allows users to find and evaluate subsets of attributes and then apply on regression models. Different views show univariate and multivariate quality metrics summary to enable the model evaluation by individual and combined attributes. Instead of testing combinations of subsets applied to models, we expose a correlation map of attributes with labels so that users can test the predictive power of attributes for each label individually as well as for multiple labels toward multinomial regression models. Strongly correlated attributes can also be detected by the interactive resources, thus allowing the user to deal with the multicollinearity problem.

The uniqueness of our approach lies in the friendly and transparent manner to create and evaluate regression models. With only a few clicks, non-expert users can gain insights and generate predictive models about their data sets. We also created a new design for the visual evaluation of multinomial LR models by employing a novel RadViz-based mapping concept.

\section{Methods}
\label{MM}

In this section, we describe the general methodology of UCReg. Figure \ref{fig:pipeline} shows the pipeline of the approach. First the analyst performs an attribute exploration and feature selection related to a target label set amongst the categorical fields of the data set. the basis for that is information on correlation between attributes and target labels as well as feature ranking. Then she builds and evaluates a Linear Regression model with the support of ROC curves and the novel Radviz-variant visualization. The analyst is free to evaluate various combinations of subsets of attributes to define the model more accurately. After the analyst has gathered a considerable amount of information and is satisfied with the selected features, he or she can create a query panorama based on a finalized LR model. Following the completion of the model, an overview tool is generated for queries, which can be saved and accessed later without the need to repeat any previous steps. When querying  a particular attribute-based profile in the tool, the analyst should provide the associated values of model's variables chosen for the Regression, and the tool will show probabilities of outcomes as well as the cases in the data set most correlated with the query. Finally, it is also possible to visualize the evolution of outcomes through streamgraphs by submitting updated attributes values, assuming the existence of attributes that vary dynamically. 

\subsection{Step 1 - Interactive Feature Selection}

After loading the data set, represented by a table of variables with numerical values for all samples, the first step in this approach is to perform an Feature Selection (FS). As mentioned earlier, LR models are sensitive to strongly correlated independent attributes. Thus, a good FS is required, aiming at selecting attributes incrementally that have potential to describe the target label.

Unlike the automatic FS algorithms, interactive selection allows the user to employ his prior knowledge as a criterion of choice. For example, in trauma EHRs, several attributes correlate with the patient's final condition, especially the trauma scores. One of the known scores is the \textit{Revised Trauma Score (RTS)} \cite{champion89revision}, which is a combination of other variables: glasgow coma scale (GCS), systolic blood pressure (SBP) and respiratory rate (RR). Therefore the user should in principle select either the three variables or RTS alone. Hence, user selection typically carries some tacit knowledge not always detected by automatic algorithms.

\begin{figure}
	\centering
	\includegraphics[width=14 cm]{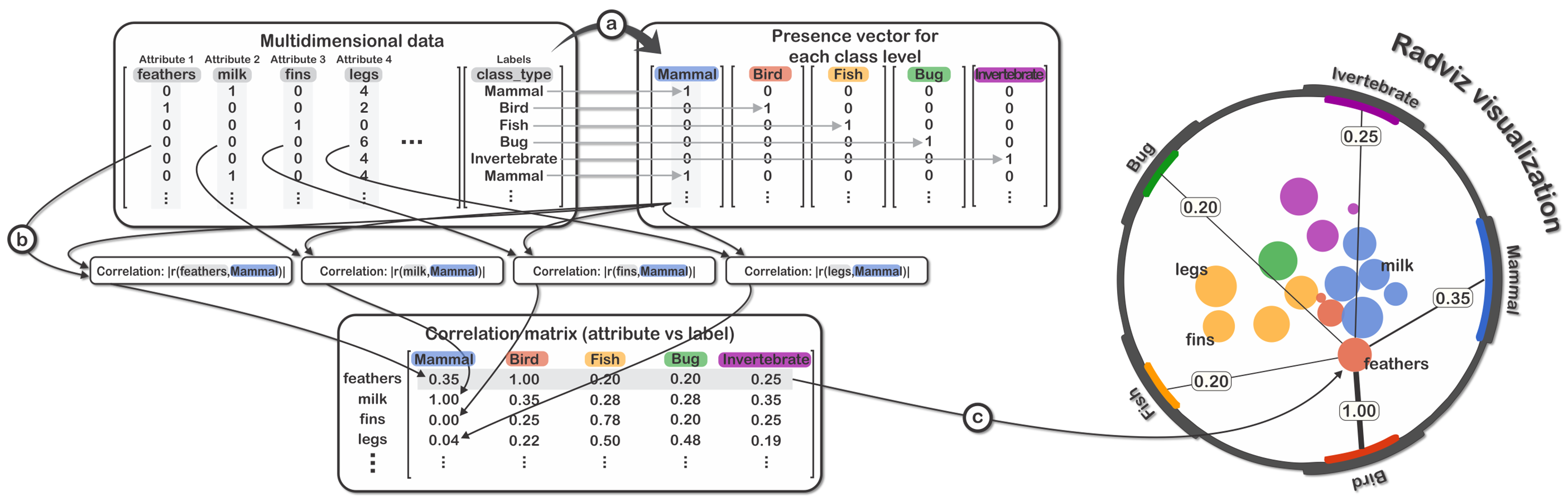}
	\caption{Graphical abstraction of the Attribute RadViz as presented in \cite{artur19radviz}, employed in the first step of UCReg. The approach shows an overview of relationships between attributes and labels in the target attribute. (a) Decomposition of labels of target attribute into presence vectors (for each label). (b) Finding correlation between data labels and other features. (c) Projection of the correlation matrix encoded into the RadViz visualization technique.}
	\label{fig:featureselection} 
\end{figure}

\subsubsection{The Attribute-RadViz}

The RadViz technique, originally conceived to represent samples, not features, has been improved and adapted to various applications. Work that adapts RadViz as support to the FS task, mostly attempt to display a summary of the attributes, highlighting the best ones from the perspective of some metric. We have adopted the Attribute RadViz, a variation of the classical RadViz technique designed to project attributes rather than data items. Its construction is simple and it is fast enough to be developed as a web-based tool that handles many attributes. For details of that tool we refer to \cite{artur19radviz}. An example is given for understanding of the FS step.

Figure \ref{fig:featureselection} shows how Attribute RadViz works. Firstly, the target attribute is decomposed into presence vectors, one for each actual label (see Figure \ref{fig:featureselection}a). The correlation calculation between attributes and labels is then performed. The resulting matrix stores correlation values for each ``attribute versus label'' combination (see Figure \ref{fig:featureselection}b). In this away, the original RadViz mapping is used to project each attribute in the visualization area under the influence of the data labels as dimensional anchors (DA). Thus, each label attracts the most correlated attributes to it (see Figure \ref{fig:featureselection}c).

Let us illustrate the positioning with an example. When constructing the correlation matrix in Figure \ref{fig:featureselection}, we can observe that the attribute ``feathers'' has a powerful correlation with the ``bird'' label and mild correlation with the others. Applying the RadViz mapping, the attribute is positioned next to the bird-DA since this anchor exerts the most significant attraction force among all DAs. Therefore, the positioning scheme constructs a cognitive map that facilitates the user exploration in the search for specific correlations by labels.

The element's sizes also give hints about the correlations relationships. Back to our example shown in Figure \ref{fig:featureselection}, the largest mapped element is the attribute ``legs''. That is because, in the context of all label versus all attributes, the most interesting attribute is actually ``legs''. However, if users are looking for strong correlations specifically for ``Fish'', they can hover the pointer over the Fish-DA (a.k.a. as focus DA), and the size of the elements becomes coded with the perspective of ``Fish'', so the ``fins'' attribute will become the largest element of the view momentarily. Additionally, users can investigate the correlation between attributes themselves. This feature is essential to avoid the selection of strongly correlated attributes. To do so, the user can hover the pointer over each one, and the sizes of the others become proportional to the correlation related to the focus DA.

The colors of the elements represent the label they are mostly correlated to, that is, they assume the same DA-label color. For instance, looking at Figure \ref{fig:featureselection}, the ``feathers'' attribute has a higher correlation with ``bird'' label, as well as ``milk'' with ``mammals'' and, ``fins'' and legs with ``fish''. Positive and negative correlations are not different from the perspective of the visual artifacts identifying correlation.

\begin{figure} [t]
	\centering
	\includegraphics[width=12 cm]{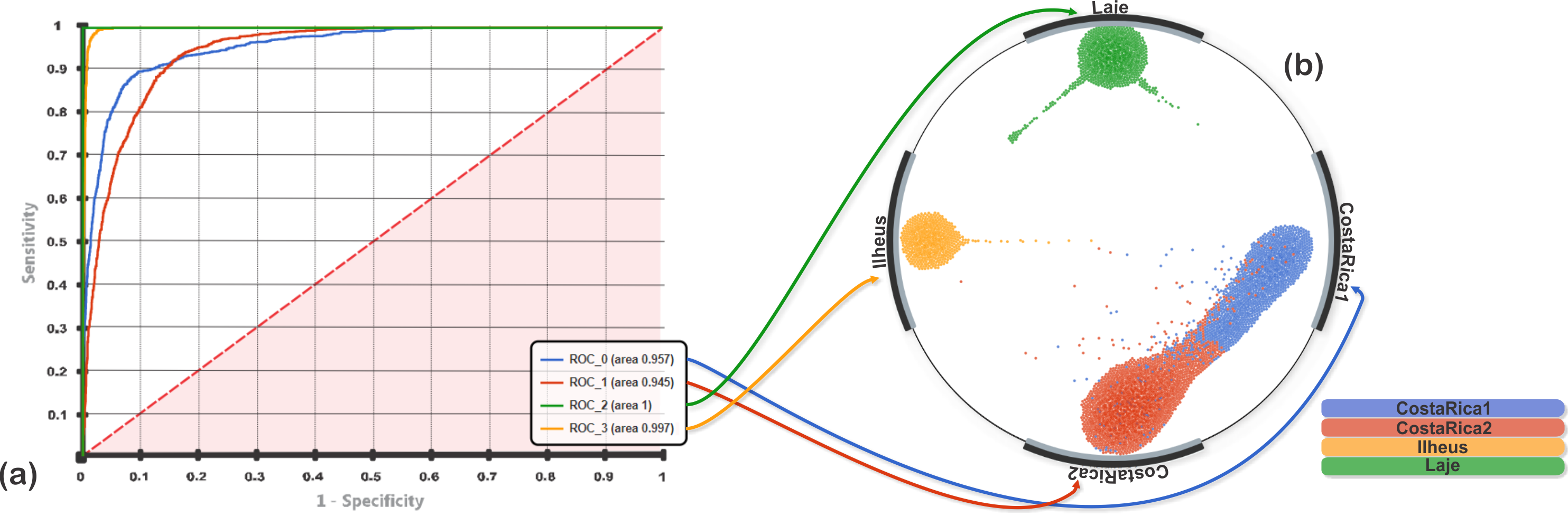}
	\caption{Distinctive alternatives evaluating the same regression model generated by the approach, in this case, the identification of acoustic data origin from four different regions. (a) ROC curves identify the individual efficiency of the binary LR as well as the ideal cut-off values for each model. (b) The LoRRViz displays by the proximity of each DA how each item is classified by the multinomial model, giving a broader view of how the model might be missing.}
	\label{fig:enhancedradviz} 
\end{figure}

\subsection{Step 2 - Logistic Regression}

Logistic regression is widely used when one wants to predict the probability of an outcome that is commonly binary. Its simplicity and elegance make it a popular solution to classification problems \cite{cox58regression}. The model is generated from observations where one or more independent attributes (discrete or continuous) determine an outcome. In our approach, we employ both classical binary LR and its generalized variant, the multinomial LR.

Binary LR is a special type of regression where a dependent attribute, which represents a binary outcome, is related to a set of independent attributes, also known as explanatory attributes. Binary LR differs from other regression types in that it does not attempt to predict a value through the linear combination of the independent attributes. It tries to predict the odds and probabilities of a given event to occur or not.

For a given number of attributes, $x_1, x_2, ..., x_n$, the probability of the response attribute is a function $y$ of the given explanatory attributes. The prediction function is defined by equation \ref{eq:logreg}.

\begin{equation}
\label{eq:logreg}
y=\sigma(w_0+w_1x_1+w_2x_2+...+w_n+x_n)
\end{equation}

Where $w_0, w_1, ..., w_n$ are the coefficients (or weights) associated with different explanatory attributes and $\sigma(z)$ is the logistic function. The logistic function is usually calculated by the sigmoid function, defined in equation \ref{eq:sigmoide}.

\begin{equation}
    \label{eq:sigmoide}
    \sigma(z)=\frac{1}{1+e^{(-z)}}
\end{equation}

We apply binary LR to generate predictive models related to isolated label values present in the dependent (target) attribute chosen by the user. However, often the user wants to examine the probability scenario between the various label values of the target attribute into unified charts. For this purpose, we also employ the multinomial LR.

The solution we adopt to solve the multinomial regression splits the problem in a set of $k-1$ binary logistic models, $k$ being the number of label values inside the dependent attribute. For each sub-problem, we find the coefficients of the model and apply in the logistic function, which is currently the softmax function given in equation \ref{eq:softmax}.

\begin{equation}
    \label{eq:softmax}
    \sigma(z)_j=\frac{e^{z_j}}{\sum_{1}^ke^{z_k}}
\end{equation}

Hence, we can provide a complete scenario of outcome probabilities to the user, both concerning the odds of isolated labels to occur and as multi-label situations. 

\subsection{Step 3 - Evaluation of Logistic Regression }

We propose to use visual and interactive means to evaluate the LR models generated in UCReg. The first one is the well-known ROC curve, which is generated as soon as the LR model is ready. The second method of evaluation is through an interactive strategy, for which we adapt the RadViz visualization technique. In the following, we describe details of implementation and the description of the interactive interface of each of these two tools.

ROC curves are widely adopted for performance analysis in classifiers \cite{gonccalves14roc}. It describes the relationship between the specificity (true negative rate) and sensibility (true positive rate) beyond the scope of a threshold assumed by some diagnostic test. A handy measure extracted from the overall performance of the classifier examined in the ROC curve is the area under the curve (AUC), and it can be interpreted as the average value of sensitivity among all possible specificity values \cite{Zhou11stat}. In our approach, along any point in the path of the ROC curve, users can consult the sensitivity, specificity, AUC, and cutoff values. Various ROC curves can be plotted simultaneously to compare the generated models, as shown in Figure \ref{fig:enhancedradviz}a. Also, the user can consult hidden information of each generated model interacting with the ROC view legends, as the overall model fit values (e.g., p-value), as well as the generated regression equation.

ROC curves are excellent for evaluating the performance of a classifier when the response is binary. Nevertheless, when the scenario includes a multi-label condition, that is, users want to hold more than two categories in the model; a multinomial LR modeling should be applied. For these cases, we can analogously evaluate the models separately by several ROC curves. However, some details of the model evaluation remain unclear, such as how the models are missing the mark.

We have developed a Logistic Regression Radial Visualization (LoRRViz) to reveal the efficiency of the multinomial regression model visually (see Figure \ref{fig:enhancedradviz}b). Each DA represents a label value and exerts attraction force according to the probability value for each item defined by the LR model. If an item is placed very close to some label-DA, it implies that the probability value of this item regarding that label is high (and low regarding the other labels). In contrast, if an item is placed equidistantly between two labels, it may mean that the probability defined by the model is the same for both labels represented by the DAs. Interactively, the user can further investigate these cases of inconsistent probabilities generated by the model, locate outliers and typical examples.

\begin{figure} [t]
	\centering
	\includegraphics[width=15 cm]{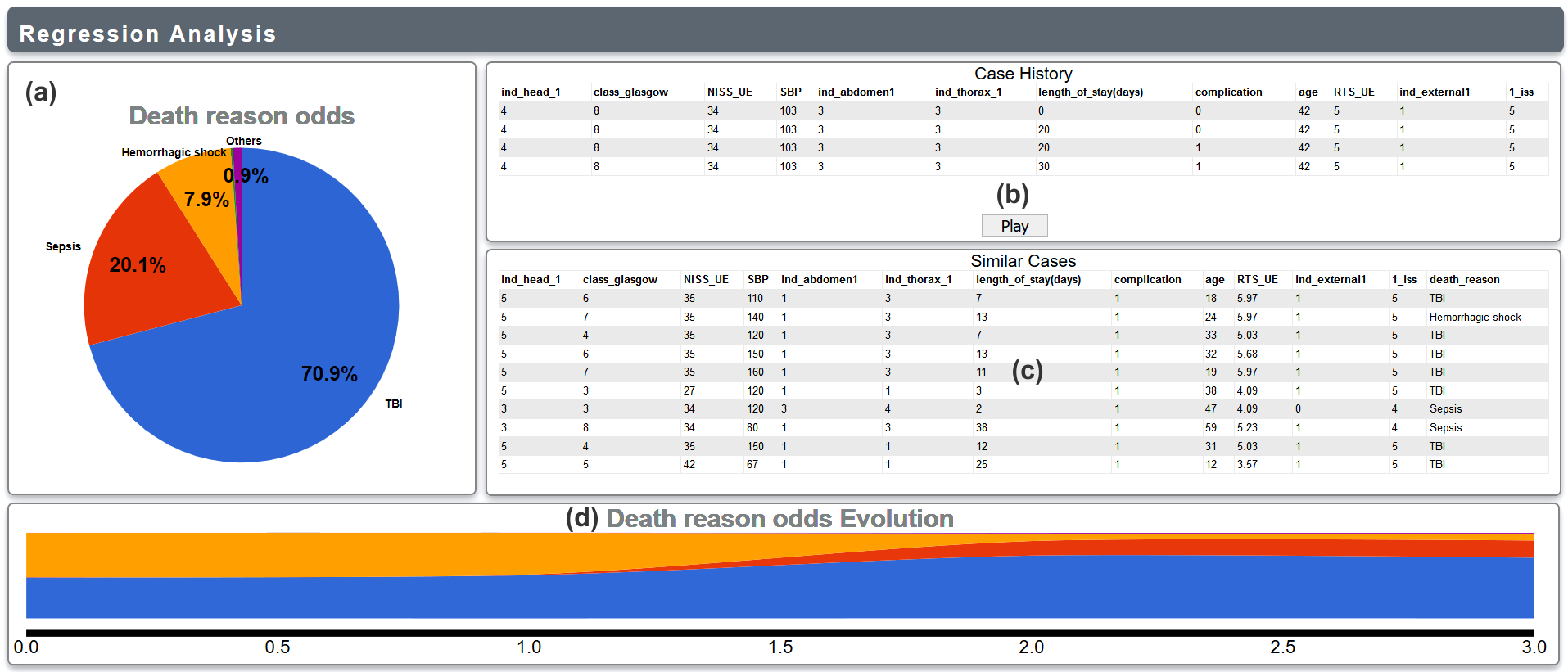}
	\caption{Query tool interface composed of four panels. (a) Pie chart presenting the probabilities calculated by the regression model for the consulted profile. (b) The history containing submitted profile states allowing the user to perform comparisons. (c) Cases most correlated with the last consulted profile. (d) Streamgraph containing all submitted states exposing trends in the evolution of probabilities.}
	\label{fig:queryinterface} 
\end{figure}

In practice, the entire matrix of $n$ items versus $k$ labels is assembled through the Equation \ref{eq:softmax} (or Equation \ref{eq:sigmoide} if only one label is chosen), resulting in a $n$ x $k$ matrix. The interpretation is that each cell holds the probability that item $x_i$ belongs to the label $l_j$. Finally, the RadViz mapping is applied, where the interactive resources available in the Attribute-RadViz such as the force scheme improvement, DAs management, and positioning distortion adjustment are also available in LoRRViz.

Figure \ref{fig:enhancedradviz} shows the evaluation of a multinomial regression model generated from an acoustic data set. The data has been collected in four different regions; their attributes have been extracted, and then we have created an LR model able to identify the origin of the samples by our approach. The color represents the actual origin, and the position represents the way in which the model classified the samples. We can note that, for the ``Ilheus'' and ``Laje'' categories, the model classified with quite a significant accuracy. However, in the ``Costa Rica 1'' and ``Costa Rica 2'' categories, we notice a considerable mix of samples, and it is up to the user to investigate whether the model needs adjustments focusing on these categories (such as new selection of attributes), or whether the samples are just intrinsically indiscernible regarding the available attributes.

\subsection{Step 4 - A Regression Query Tool}
\label{aregquerytool}

UCReg proposes and implements the generation of regression models according to the user's interest in predicting single or multiple events. Also, we include resources to enable model evaluation and reveal potential adjustment demands. However, aiming for an all-around solution, we have developed a query tool to take advantage of the previously generated learned data for later queries related to any desired profile.

The tool usage is quite simple; the user restores previously saved regression data and then inserts the information about a profile that he or she wants to query; hence, the query tool will return the probabilities of this profile belonging to any of the labels. He or she can also submit several states of this profile; in this way, the tool will expose evolution trend patterns. For example, in the case of a patient that varies his clinical condition over time, the user can submit his states and observe the variation of probabilities of outcome trends exposed by the model.

Figure \ref{fig:queryinterface} shows the query tool interface. There are four panels per generated model. The first one (see Figure \ref{fig:queryinterface}a) presents a pie chart with the probability scenario of the current consulted profile. The second (see Figure \ref{fig:queryinterface}b) presents the history of submitted states. The next panel (see Figure \ref{fig:queryinterface}c) shows the most correlated cases with the last consulted profile. To make this panel available, the user must, in addition to opening the regression model file, to load the original data set. Finally, the last panel (see Figure \ref{fig:queryinterface}d) shows, through a streamgraph, the evolution of the probabilities according to the profile changes submitted by the user. This panel is useful when the analyzed object has attributes that change dynamically over time and reveals trends in the probability scenario.

\begin{figure}[t]
    \centering 
    \begin{subfigure}[b]{0.25\textwidth}
    \includegraphics[height=100px]{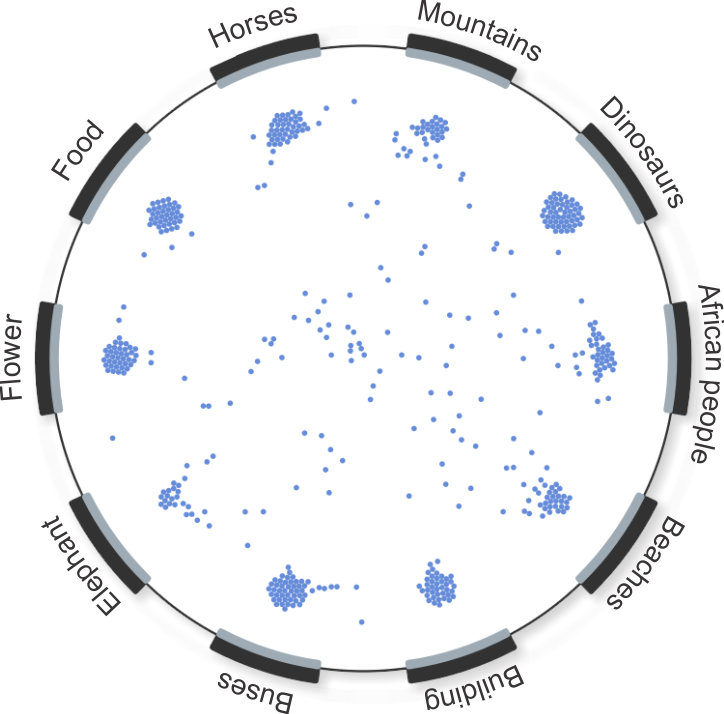}
    \subcaption{}
    \label{fig:radvizregression:a}
    \end{subfigure}
    \qquad \qquad
    \begin{subfigure}[b]{0.25\textwidth}
    \includegraphics[height=100px]{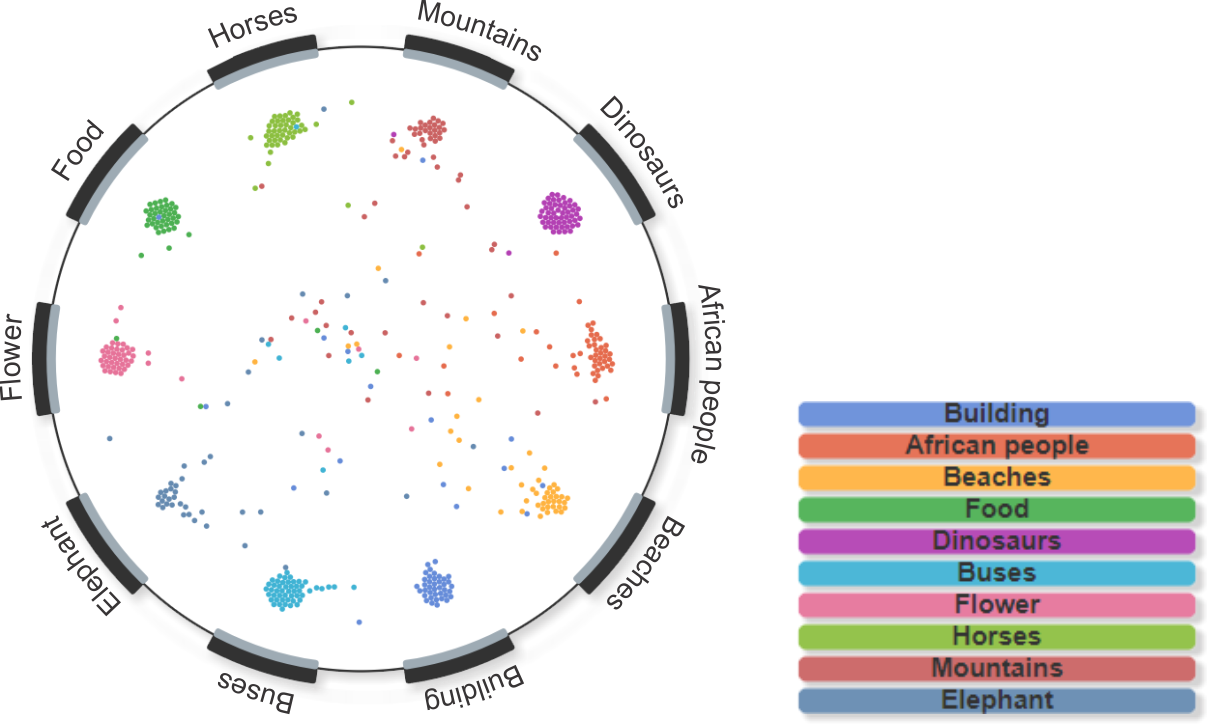}
    \subcaption{}\label{fig:radvizregression:b}
    \end{subfigure}
    \caption{Viewing estimated results by the multinomial LR model with the test partition (half of all samples) of the Corel data set. In addition to individual profile queries, users can query entire data sets through CSV files, where the tool seeks the attributes involved in the chosen regression models. The visualization can be rendered (a) without the label information or (b) applying the labels to encode the colors of mapped elements.}
   \label{fig:radvizregression}
\end{figure} 

The tool also allows multiple-model queries. For example, in a trauma data set, we can find more than one target attribute in which we would like to create prediction models, such as the ``condition of discharge'' and ``cause of death'' attributes. Hence, we can imagine three scenarios that can be queried simultaneously: ``probability of survival'', ``condition of discharge (in case of survival)'', and ``cause of death (in case of obit)''. Thus, the user can set up a more comprehensive odds scenario provided by the regression models generated from the same data set.

There is also the possibility of querying not only individual profiles but entire lists of items. As soon as the data is ready, the tool scans for the selected explanatory attributes of the built regression models, and them an overview of the classification is displayed through the LoRRViz. Optionally the user can encode the color of the mapped elements based on some chosen attribute (usually the one containing the labels); this is useful for visualizing the efficiency of classification generated over supervised data (see Figure \ref{fig:radvizregression}).

\section{Results}

\subsection{Interactive Regression Explorer Prototype}

We have implemented UCReg as a web-based tool predominantly encoded in Javascript. Also, we use the D3js library to handle visual elements and HTML with CSS for the construction of the front-end available to the user. The regression module is also encoded in Javascript, and it is originally obtained from The Interactive Statistical Pages (\url{http://statpages.info/}). The prototype related to this work and its complete source code are made freely available at \url{https://github.com/erasmoartur/ucreg}.

\begin{figure}
	\centering
	\includegraphics[width=13 cm]{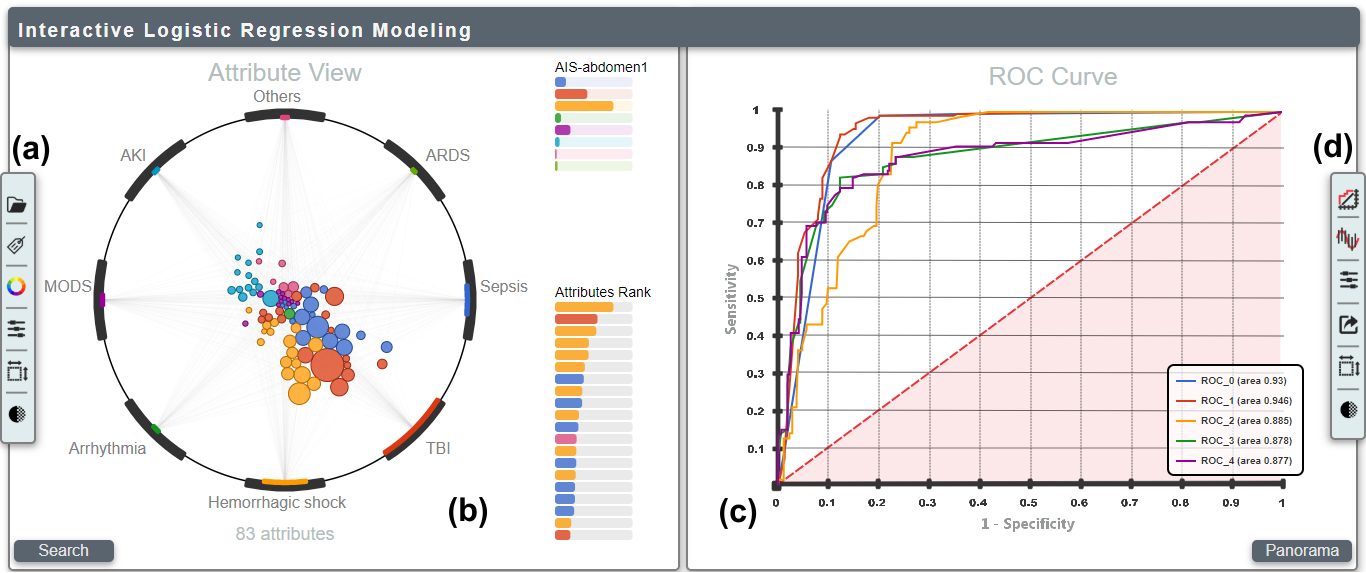}
	\caption{Snapshot of UCReg's prototype main interface. (a) The attribute view control panel. (b) Visualization of attributes through RadViz, which constructs a cognitive map showing the relevant attributes for each label value of the data set. (c) Panel with means of evaluation and review of generated regression models. (d) Control panel where the user defines the evaluation mode and changes settings about them.}
	\label{fig:maininterface} 
\end{figure}

Figure \ref{fig:maininterface} presents the main interface of the prototype. On the left side, a control panel (see Figure \ref{fig:maininterface}a) allows the user to open the data set, define the target attribute, and change visual attributes panorama settings. The right side contemplates the evaluation and check mechanisms for generated regression models (see Figure \ref{fig:maininterface}c), employing either  LoRRViz or ROC curves. The user selects the evaluation mode from the right control panel (see Figure \ref{fig:maininterface}d), as well as the settings for that view. 

In the following section, we present a step-by-step procedure applying our prototype to create and explore LR models. The experiment sequence follows the pipeline shown in Figure \ref{fig:pipeline}, focusing on a practical process over synthetic and real data sets.

\subsection{Employing UCReg}

We present four usage scenarios investigating EHR data sets to reach different purposes. In the first two scenarios, we employ synthetic trauma records data set based on a real data. Firstly, we load the data set to explore combinations of relevant attributes to generate LR models and further confront these models with well-known trauma scores.  Then we show how to take advantage of the generated models to build a query tool and also how to apply it to gain insights into the presented predictions. In the other two scenarios, we report a brief analysis and the construction of LR models for a data set referring to the novel COVID-19 disease.

\subsubsection{Scenario One: Predicting Mortality with Trauma Scores}

Trauma scores represent an attempt to characterize and document traumatic injuries levels \cite{junior99indices}. It can help in predicting the outcome of the patient and aid in the triage of the trauma patients. We compare the effectiveness of well-known trauma scores related to the patient risk of morbidity. The scores are \textit{Trauma Revised Injury Severity Score (TRISS)}, \textit{Revised Trauma Score (RTS)}, and \textit{Injury Severity Score (ISS)}; then we create and test variations of these scores in an attempt to check and increase the prediction efficiency.

In this scenario, we use a synthetic trauma data set based on real data. We designed a genetic algorithm (GA) with fitness function corresponding to the person's correlation coefficient between the individual and a sample randomly picked from the real data set. At the end of each GA process, the newly generated individual (the fittest of the last generation) is accepted for the synthetic data set if its correlation with the real chosen sample is higher than 0.95, ensuring its similarity to real data. The original data set has 21,294 records with 145 attributes collected in the 9-year interval (2006 to 2014). It includes patient profile data, trauma event information, clinical tests and observations, and calculated trauma scores. The synthetic data set has 10,000 records with 84 attributes (some attributes has been already filtered in this step). Although it is a similar data set with a real one, we assume it is enough to present the features of our approach.

Our test/training ratio corresponds to 0.2; the metric for comparison is the AUC; which is automatically displayed by the approach as soon as the model are created. Our intention here is not to develop a new improved score; we would like to show the usefulness of the approach for analysts to investigate and test hypothesis creating their own regression-based models in a practical and rapid manner.

\begin{figure}
	\centering
	\includegraphics[width=12 cm]{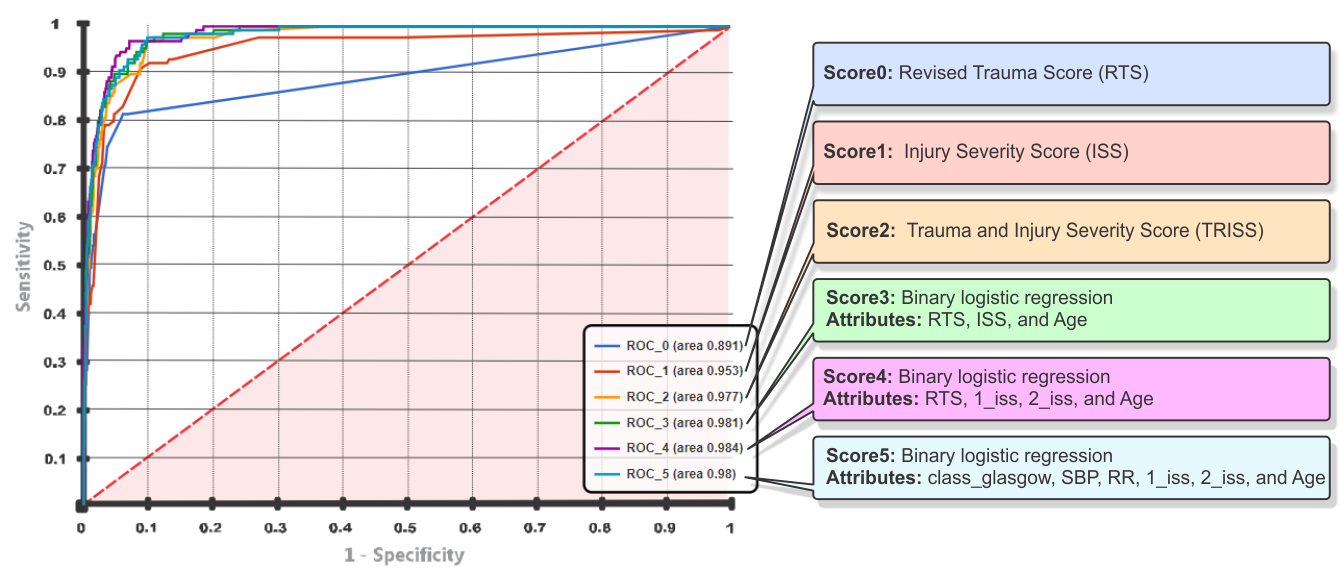}
	\caption{A comparison of the performance between well-known trauma scores and generated regression models with the training data set. Score 4 was the most accurate for this data, where its concept is derived from the traditional TRISS, but it includes the values of the two worst injury segments instead of the ISS.}
	\label{fig:scorestest} 
\end{figure}

Figure \ref{fig:scorestest} shows the ROC curves generated by applying the scores to predict mortality. The ROC curves of \textit{RTS}, \textit{ISS}, and \textit{TRISS} scores are rendered directly as their values are present in the source data set. Then, we have generated scores 3, 4, and 5. The score 3 employs the \textit{TRISS} concept, which is essentially an LR involving \textit{RTS}, \textit{ISS}, and age; the efficiency increase (AUC equal to $0.981$ against $0.977$ of the original \textit{TRISS}) is due to the new regression training, which adjusts its coefficients according to the local reality of the data under analysis. Domingues et al. \cite{de17performance} discuss the applicability of \textit{TRISS} in different contexts explaining this effect.

Scores 4 and 5 are hypothesis tests that we raised, where we may improve the \textit{TRISS} score by changing attributes of its original equation (\textit{RTS}, \textit{ISS}, and \textit{age}) by their primitive ones. In Score 4 we replaced the \textit{ISS} values -- which represent the sum of squares of the \textit{Abbreviated Injury Scale (AIS)} values of the three segments with the most severe injuries -- by the two most severe injury segments, and, in this data set, the efficiency of the score has improved (with AUC value of $0.984$). The same idea has been extended to Score 5, where instead of \textit{RTS} -- which is the weighted sum of \textit{GCS}, \textit{SBP}, and \textit{RR} -- we insert those values directly to the regression model. However, the result is not so good compared to the previous score.

Scores 4 and 5 are hypothesis tests that we raised, where we may improve the \textit{TRISS} score by changing attributes of its original equation, which is an LR with \textit{RTS}, \textit{ISS}, and age, by their primitive ones. In score 4, we replaced the \textit{ISS} values -- which represent the sum of squares of the abbreviated injury scale (\textit{AIS}) values of the three segments with the most severe injuries -- by the two most severe injury segments. Hence, the new LR model contains \textit{RTS}, \textit{AIS-1}, \textit{AIS-2}, and age. Therefore, in this data set, the efficiency of the score has improved (with an AUC value of $0.984$). The same idea has been extended to Score 5, where instead of \textit{RTS} -- which is the weighted sum of \textit{GCS}, \textit{SBP}, and \textit{RR} -- we insert those values directly to the regression model. However, no improvement noted in this score compared to the previous ones (AUC value of $0.980$).

This scenario shows how the agile creation and evaluation of regression models enable analysts to raise and test hypotheses about data efficiently. Additionally, they can model scores for prediction of events of interest encoded into categorical or numerical attributes in their own data sets.

\subsubsection{Scenario Two: Building a Prediction Interface for Trauma Events}

In this scenario, we present how to take advantage of the knowledge gained from the performed tests and then build a regression model for later queries. We follow each step described in Figure \ref{fig:pipeline}, showing how the resources available in the approach allow users to create a query mechanism quickly and effectively.

\paragraph{Feature Selection Step}

In the FS step, we try to choose relevant attributes for each label that we would like to predict. Thus, we must interact with the DAs (hovering the pointer) to discover correlated attributes and then generate its prediction model.  We want to construct a scenario that returns odds of survival, discharge condition (in case of survival), and cause of death (in case of death). We have picked the most correlated attributes for each label, for example, toward the ``hemorrhagic shock'' label, the highest correlated ones are the ``AIS-abdomen'', ``AIS-thorax'', and ``blood pressure''.

\begin{figure}
	\centering
	\includegraphics[width=13 cm]{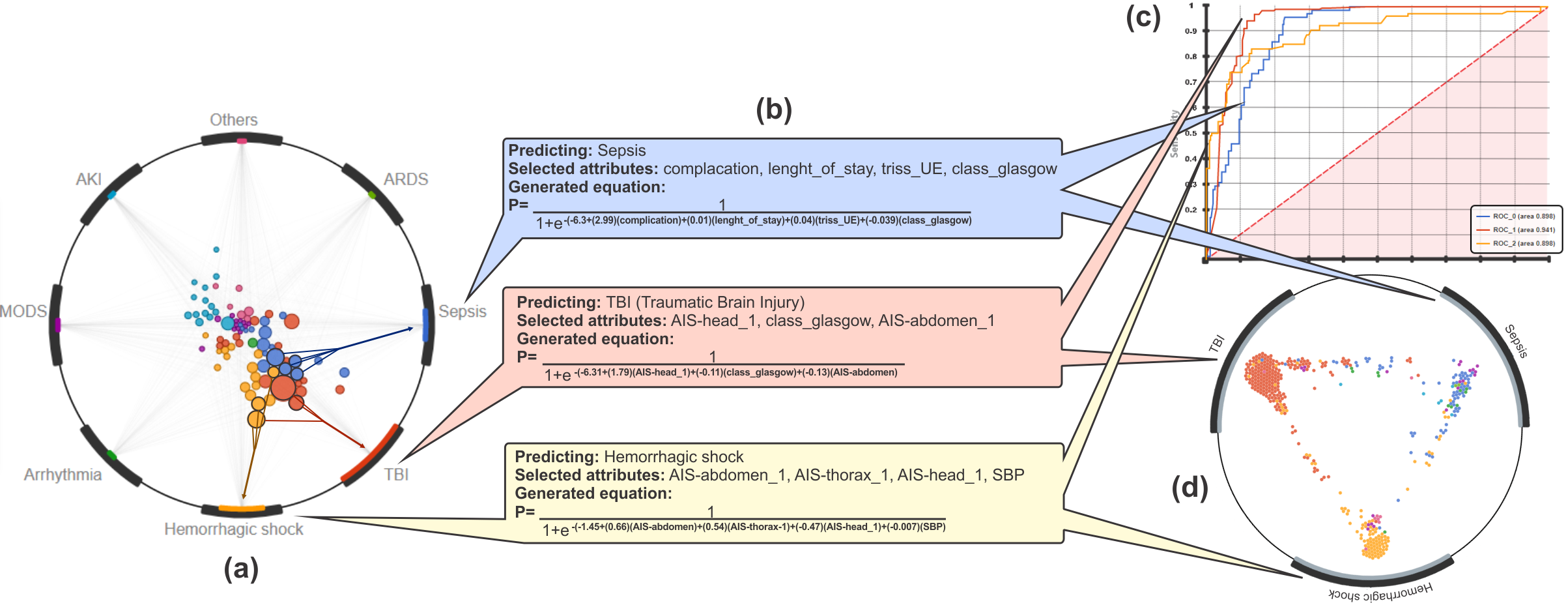}
	\caption{Generating and testing binary LR models. Before creating the definitive model for further queries, we evaluated the performance of the highest correlated attributes for each label through ROC curves. (a) Selected attributes for the labels Sepsis, TBI, and Hemorrhagic shock. (b) Logistic model created after selecting attributes and clicking over the desired DA. (c) The isolated model evaluation through ROC curves. (d) Evaluating the model by the LoRRViz.}
	\label{fig:rocexamples} 
\end{figure}

\paragraph{\textbf{Evaluating the Logistic Regression}}

Selecting attributes and choosing a target label (by clicking on the DA) generates an LR model. Then its performance as a label predictor under the already loaded data is immediately presented through ROC curves on the right side of the prototype. Figure \ref{fig:rocexamples}c shows ROC curves plotted by the generated models for prediction of labels ``Sepsis'', ``TBI'', and ``Hemorrhagic shock''.

Since we want a scenario with multinomial regression models, we can use the LoRRViz to evaluate the generated models. To do so, in the right control panel, we chose LoRRViz. Figure \ref{fig:rocexamples}d shows the classification of the models generated so far, including its inconsistencies; the analyst can investigate such cases and proceed, if necessary, with adjustments in the model.

\paragraph{\textbf{Multinomial Logistic Regression Set Up}}

Once we gain a better understanding of the data set, and we test the predictive power of the attribute's subsets related to the labels in regression models, we can then generate a definitive and generic model. This model should allow future queries without the need to perform again the steps described previously.

\begin{figure} [t]
	\centering
	\includegraphics[width=15 cm]{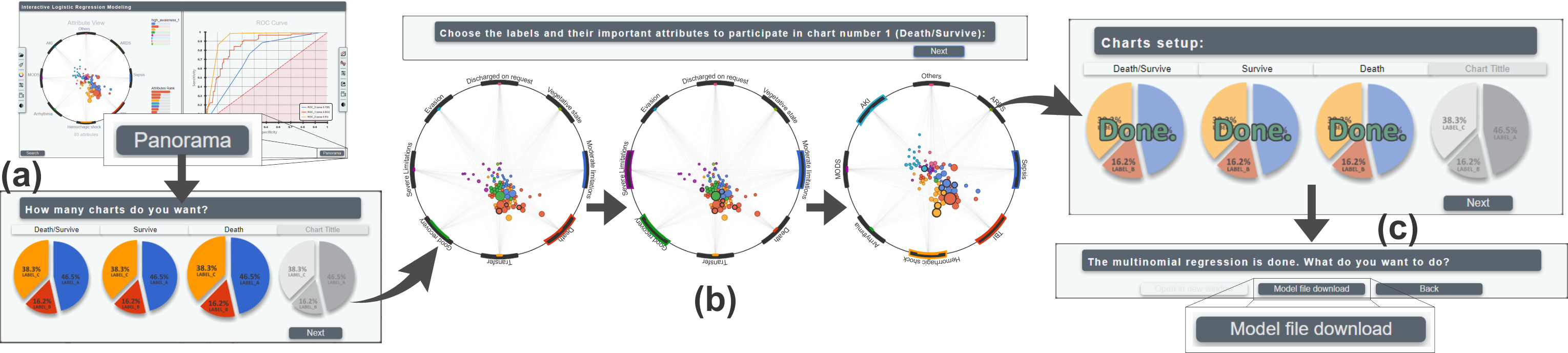}
	\caption{Step by step illustration when generating the definitive model for this scenario. (a) After clicking in ``Panorama'', the prototype prompts for the number of models/graphs to be created (in this case three), and their respective names. (b) Then, we enter the labels and attributes for each model. (c) Finally, the prototype returns the file with regression learning.}
	\label{fig:querysetup} 
\end{figure}

Inside the prototype, we click on the button ``panorama'' and it requests the number of graphs we want and their titles, as shown in Figure \ref{fig:querysetup}a. Users can choose more than one graph to make the tool able to display the probabilities of more than one outcome simultaneously, creating a more comprehensive overview. For example, in our case, we chose three graphs; the first presenting the probabilities of survival versus death; the second presenting the most probable discharge condition in case of survival; and the third presenting the most likely cause of death if the patient does not survive.

Then, we determine which labels and attributes will be part of each model (see Figure \ref{fig:querysetup}b). When only one label is chosen for some model, it is assumed that the user wants the binary LR scenario, where the odds are given for the label's chance to occur or not. Here we choose the label death, that is death versus all, consequently death versus survival. The selected attributes are ``RTS'', ``Age'', and ``ISS''. We click next to proceed to the second graph set up. Then we exclude the death label (only survival labels left) and choose the following labels that represent a discharge condition: ``Good recovery'', ``Moderate limitations'', and ``Severe Limitations.'' The attributes selected here are ``RTS'', ``Age'', ``ISS'', ``AIS-head\_1'', ``surgery'', ``previous\_pathology'', and ``complication''.

For the last graph, we change the target attribute to ``cause death'', and we chose the following labels: ``TBI'', ``Hemorrhagic shock'', ``Sepsis'', and ``AKI''. Then, the attributes ``AIS-head\_1'', ``complication'', ``length\_of\_stay'', ``AIS-abdomen'', ``AIS-thorax'', ``AIS-external\_1'', ``SBP'', and ``RTS'' are selected. Finally, we click next to proceed with the creation of the model and download of the file containing the regression learning data (see Figure \ref{fig:querysetup}c).

\paragraph{\textbf{Querying the Models}}

After the generation of the models, we can perform queries to check the odds of the desired profiles. Inside the tool, we put the pointer on the top bar to make the input menu appear. Inside it, we insert the previously downloaded learning data and, optionally, we can add the original data so that the tool is also able to show the most correlated cases with the queried profile, this is useful for the user to verify if the results indicated by the models are similar to the cases inside the original data set.

\begin{figure}
	\centering
	\includegraphics[width=13 cm]{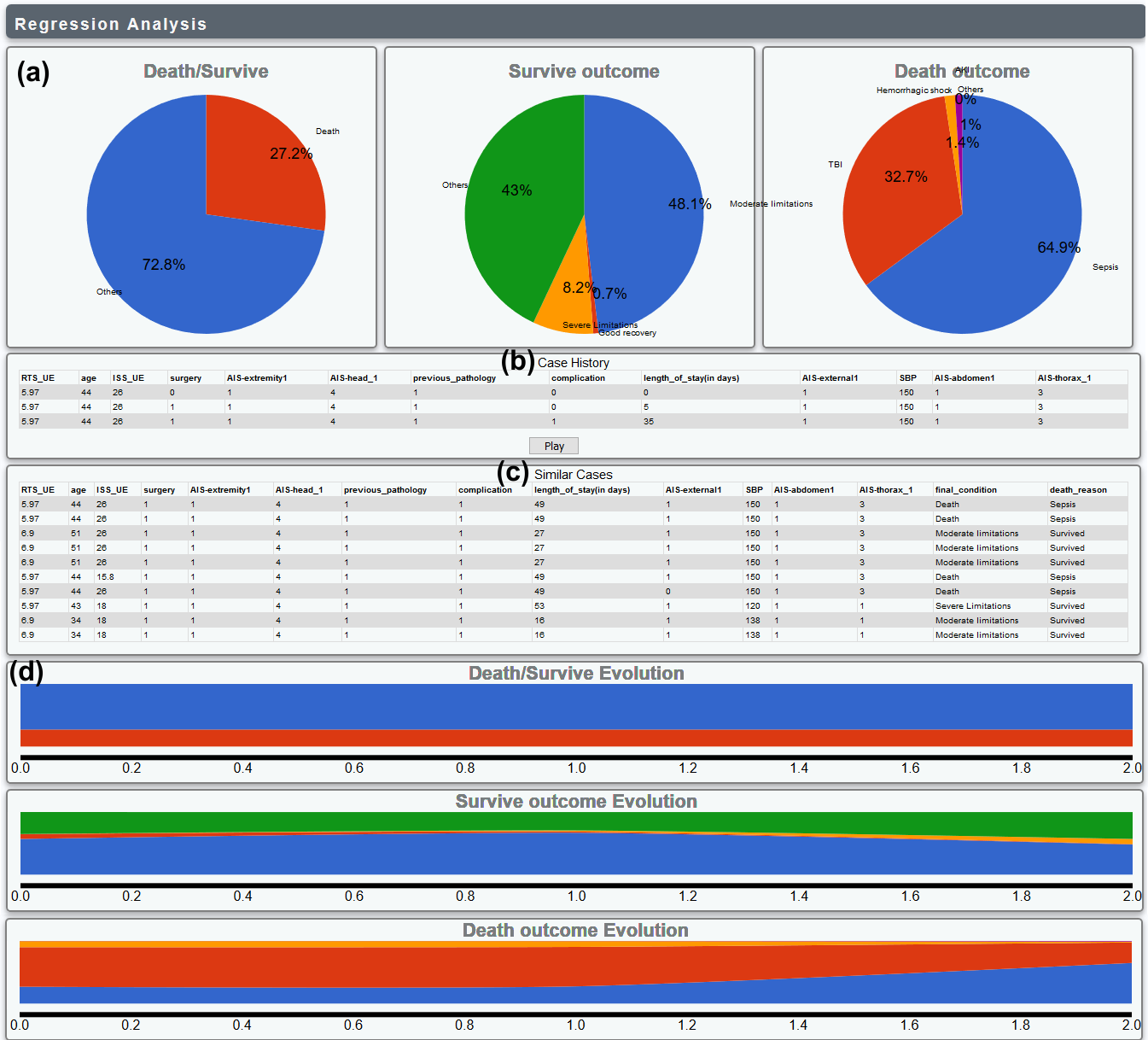}
	\caption{The query tool loaded with the regression learning data previously described. (a) The graphs are showing the calculated odds for the queried profile. (b) The submitted profiles (generally over time) for analysis of outcomes trends. (c) List containing the most similar cases to the last submitted profile. (d) The streamgraphs are displaying trends related to the states submitted in (b).}
	\label{fig:queryexample} 
\end{figure}

Since we have created three models, when we restore the file containing the learning data, the tool requests values of the previously selected attributes of these models to set up the profile to be queried. As we also inserted the original data set, the average values of each attribute are automatically filled inside each input box. We then simulate a situation witch a patient is admitted in the hospital in a particular condition, but his or her attributes change dynamically over time, so we submitted three more condition states of the patient. Figure \ref{fig:queryexample}a shows the odds for the last queried state. Figure \ref{fig:queryexample}b shows all submitted sates, where we simulate that the patient underwent a surgical procedure, changed the state of complication, and naturally spend more few weeks hospitalized. In Figure \ref{fig:queryexample}c, a list presents the cases most correlated with the last submitted state. Finally, in Figure \ref{fig:queryexample}d, we see the streamgraph of the submitted states for each model. In the third streamgraph, it is possible to observe an increase in the chances of Sepsis, and it may represent a tendency for the patient to suffer from this condition, which could serve as support to shape the decisions of the healthcare team.

This scenario has shown that users can apply the tool to jointly identify outcomes as well as trends in the evolution of the characteristics of the object of interest. This feature assists the analyst not only in predicting current status but also in supporting complex decision-making situations. This aspect is difficult to find with the previous tools designed to explore regression models.

\subsubsection{Scenario Three: Investigating the Novel COVID-19 Disease}

An ongoing global outbreak of a virus, initially identified in the region of Wuhan, China, at the end of 2019, has spread and generated turbulence never seen before in modern human history. The World Health Organization (WHO) has officially named the virus as SARS-CoV-2, which means severe acute respiratory syndrome coronavirus 2.  The infection caused by the virus has been named as coronavirus disease 2019 (COVID-19) also by the WHO \cite{Feng19Corona}. Since the spread of COVID-19 outside Chinese borders, the global scientific community has been dedicating enormous effort to understand the virus and the infection that it causes.

In this scenario, we have employed one of the freely available COVID-19 data sets to investigate the correlations of attributes and generate insights about the data. The focus is to understand how the onset symptoms, patient profile information, and chronic disease historic relate to the outcome of the treatment. Then, we built an LR model for the prediction of severe cases, and we compare results with the literature that outlines statistics of different patient profiles.

The data set is available from Kaggle\footnote{https://www.kaggle.com/sudalairajkumar/novel-corona-virus-2019-dataset} and obtained on March 31, 2020. It contains more than 33,000 reported cases. However, given our objective in this scenario, we have filtered out cases that do not specify age and symptoms. It reduced the data set to 1,586 occurrences, of which 56 cases resulted in death, and the others were either still active or cured until the date of download. We also decompose the symptoms attribute, which describes the set of symptoms for each report, into individual attributes for each of the most frequent symptoms.

\begin{figure}
    \centering 
    \begin{subfigure}[b]{0.6\textwidth} 
        \includegraphics[width=\textwidth]{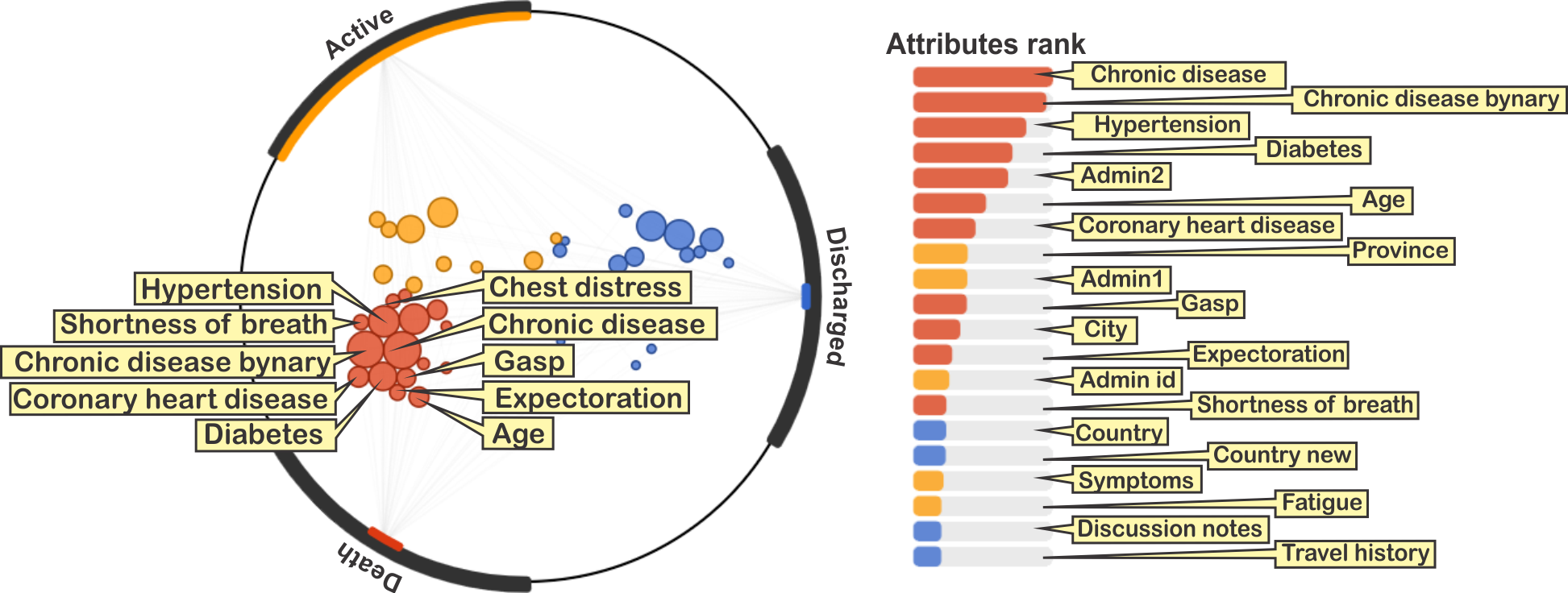} 
        \subcaption{Initial attribute view} 
        \label{fig:covid:attrs} 
    \end{subfigure}
    \begin{subfigure}[b]{0.3\textwidth} 
        \includegraphics[width=\textwidth]{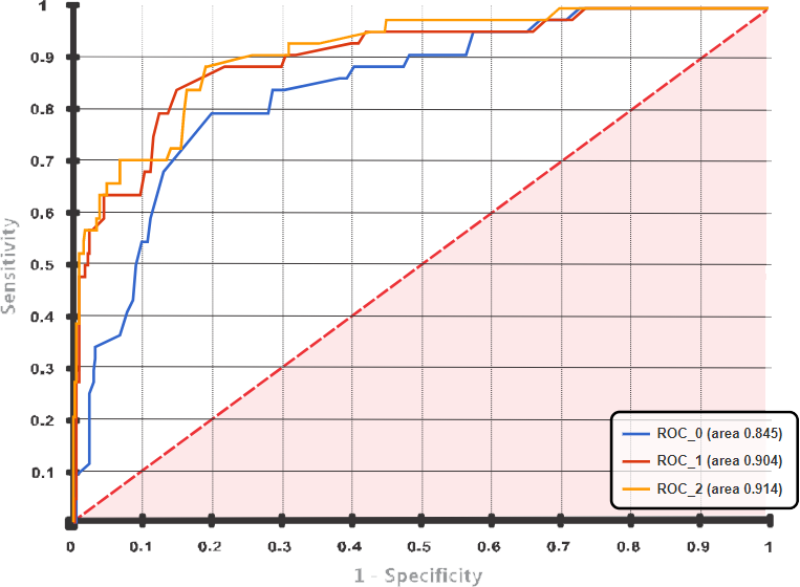} 
        \subcaption{ROC curves} 
        \label{fig:covid:rocs} 
    \end{subfigure}
    \caption[Investigating the COVID-19 data set.]{Investigating the COVID-19 data set. (a) Attribute-RadViz shows correlations to the outcome labels; in addition, users can consult the dynamic rank. (b) Three binary LR models have been created; the blue one with only the ``age'' as an explainable attribute, the red one adds the set of attributes related to chronic diseases, and the yellow one includes all the previous ones plus the set of attributes related to the symptoms.} 
\end{figure}

Initially, we want to understand the correlations of attributes in relation to the results of treatments; hence, we chose the ``outcome'' attribute as the target attribute to start the tool. Figure \ref{fig:covid:attrs} shows the initial rendering. We can quickly observe a group of attributes (the red ones) with potential to present high correlations with cases of death.

Investigating further, we realize our first relevant observation regarding this data set; the attributes that represent the history of chronic diseases have a significant correlation with cases of death, such as hypertension with a 0.38 estimated coefficient. On the other hand, the attributes that represent the initial symptoms of the disease have a medium-to-low or null correlation, even the symptoms considered more severe, such as shortness of breath, which has a 0.19 estimated coefficient.

Given the findings about the correlations of attributes, we have generated three LR models. The first model predicts death with only the ``age'' information. The second one contains the attributes: ``Chronic disease binary'', ``Hypertension'', ``Diabetes'', ``Coronary heart disease'', and ``age''. The last model includes all the attributes of the previous one plus the attributes that represent the most severe (for this data set) symptoms of the disease, which are: ``Gasp'', ``Expectoration'', ``Shortness of breath'', ``Fatigue'', and ``Chest distress''. Figure \ref{fig:covid:rocs} shows the ROC curves generated by the models, wherewith only age information, the model has shown relevant predictive potential. The inclusion of attributes related to chronic diseases has shown a significant increase in the prediction capacity; however, the further inclusion of the symptoms attributes produces only a modest increase in the model's efficiency for this data set. It emphasizes our first observation, in which the attributes related to the history of chronic diseases have higher predictive power in comparison with attributes that represent symptoms of the disease.

Although most of the  correlations found correspond to the patient's history and present state, other significant correlations are also exposed by the tool such as ``city'', ``province'', and extra attributes related to the place of treatment. Factors as the local demographic characteristics, climatic condition, and medical care quality may generate some correlation of death to the geo-location of the patient \cite{Wang20hightemp}. In general, it is up to the analyst to investigate the validity of these findings. In the context of this scenario, the data set has been initially collected at the peak of the pandemic event in China, so most of the death cases come from the Wuhan region, but the data set includes many other active cases from different areas and countries, such as Japan and Italy. This condition generates a false correlation of death with the patient's geo-location. Circumstances where hidden confounding attributes (such as the length of contagion of a region) create spurious correlations often can be detected by analysts, which highlights the importance of tools that insert the human into the process.

\begin{table}[]
\caption[The estimated mortality rate caused by the COVID-19 disease distributed by age groups.]{\label{tab:covidodds} The estimated mortality rate caused by the COVID-19 disease distributed by age groups. The first record comprises the values determined by the analysis in \cite{surveillances20epi}. In the following, we have estimated rates simulating the absence and presence of comorbidities and severe symptoms.}
\centering
\scalebox{.8}{
\begin{tabular}{lccccccccc}
                        \hline
                      & \multicolumn{9}{c}{Age groups}                                                  \\
                      & \multicolumn{9}{c}{rate, \%}                                                    \\
                      \cline{2-10}
                      & 0-9 & 10-19 & 20-29 & 30-39 & 40-49 & 50-59 & 60-69 & 70-79 & $\geq$ 80 \\ \hline
Surveillances \cite{surveillances20epi}          & --  & 0.2   & 0.2   & 0.2   & 0.4   & 1.3   & 3.6   & 8.0     & 14.8              \\
Our LR model              & --  & 0.1   & 0.1   & 0.3   & 0.6   & 1.3   & 2.9   & 6.3   & 13.1              \\
Hypertension=1        & --  & 0.1   & 0.1   & 0.3   & 0.8   & 1.8   & 4.2   & 9.5   & 19.8              \\
Diabetes=1            & 0.2 & 0.5   & 1.1   & 2.3   & 5.1   & 10.7  & 21.1  & 37.3  & 57.0              \\
Shortness of breath=1 & 0.1 & 0.1   & 0.3   & 0.7   & 1.5   & 3.3   & 7.2   & 14.8  & 28.1              \\
Cough=1               & 0.1 & 0.1   & 0.3   & 0.6   & 1.4   & 3.0   & 6.6   & 13.6  & 26.1              \\
Chest Distress=1      & 0.1 & 0.2   & 0.4   & 0.9   & 1.9   & 4.3   & 9.0   & 18.2  & 33.1    \\ \hline          
\end{tabular}}
\end{table}

Inside the query tool, after building the LR model, we can observe the prediction of several profiles according to what the model learned from the data set. For comparison purposes, we attached results from a reference work in the analysis of reported cases collected in China on February 11, 2020 and presented by \cite{surveillances20epi}. Table \ref{tab:covidodds} shows the acquired rates by age group. The rates exposed by the annexed work are very similar to those delivered by the LR model when ignoring comorbidities and severe symptoms. We also attached results by simulating the presence of two of the most frequent chronic diseases in the data set, as well as by simulating the presence of three different symptoms. All data we worked in this scenario, as well as the learning files, are freely available on GitHub \footnote{\url{https://github.com/erasmoartur/ucreg}}. If the reader is interested in simulating other situations, he or she can download the tool and load it on any web browser (we recommend Chrome or Firefox); there is no need for any installation process.

Much desired information is unavailable in this COVID-19 data set; general data about the patient (such as smoke habits) and data about the provided treatment (such as the adopted drug administration) could substantially expand the investigation. However, this scenario has shown that the approach is already capable of generating relevant insights and supporting possible decision-making tasks based on predictions using the resulting FS.

\subsubsection{Scenario Four: COVID-19 in Developing Countries}

Despite the relative current established situation of part of the world over the COVID-19 pandemic, the problem is far from being resolved or controlled in the vast majority of developing countries. Therefore, as of the date of submission of this document, many of these countries are still in the high stages of contagious and accumulating deaths.

\begin{table}[]
\caption{\label{tab:covidICU} The estimation of the potential need for ICU beds for a  patient according to the logistic regression build from a COVID-19 data set from S\~ao Paulo.}
\centering
\scalebox{.9}{
\begin{tabular}{ccc}
\hline
Risk factors & Low saturation=0 & Low saturation=1 \\
n            & rate, \%         & rate, \%         \\ \hline
0            & 18.8             & 34.1             \\
1            & 21.9             & 38.5             \\
2            & 25.4             & 43.2             \\
3            & 29.2             & 48.0             \\
4            & 33.4             & 52.8             \\
5            & 37.8             & 57.5             \\ \hline
\end{tabular}}
\end{table}

In this scenario, we are investigating how the tool can provide information that helps the healthcare team to make its decisions. Two central problems persist in these countries due to resource constraints: testing capacity and availability of intensive care unit (ICU). On top of these problems, we have build a query model to give the estimates of results assuming the absence of these resources.

The estimation of the potential need for ICU bed for the patient is especially interesting due to developing countries' vulnerabilities. In Brazil, for example, only 9.8\% of cities have ICU beds, according to the Brazilian national register of health facilities (CNES)\footnote{\url{http://cnes.datasus.gov.br/}}. Still, the country's system total capacity is adequate to serve the entire population (the country has approximately 35 thousand ICUs). So this issue is a matter of imbalance in the geographic distribution of resources. Therefore, the early detection of the need for ICU beds is advantageous, given the necessity to move the patient to the nearest city with a medical center including available beds.

The data set for this scenario comes from the São Paulo State Statistics Agency (SEADE) and it is available in GitHub\footnote{\url{https://github.com/seade-R/dados-covid-sp}}. SEADE maintains a panel of data on coronavirus-related cases and deaths in the State of São Paulo - Brazil, based on official data from the São Paulo State Department of Health (SES). The data includes a description of approximately 107 thousand cases, including the patient's profile, symptoms, comorbidities, risk factors, and disease outcome.

In our investigation about the correlations among cases that required admission to the ICU, we have noticed a relevant correlation only with one symptom present in the data set: low saturation. Regarding risk factors, we found small correlations, but when combined, the regression's successes increase. Then we have derived a new attribute to the data set, called risk factors, which contains the cumulative risk factors of the patient, which can be: puerperal woman, heart disease, chronic hematology, down syndrome, liver disease, asthma, diabetes, neurological disease, pneumopathy, immunosuppression, chronic kidney disease, and obesity. Thus, those attributes have been chosen for the regression that resulted in a model with an AUC of $0.63$. The results are shown in Table \ref{tab:covidICU}; it is possible to observe that the presentation of low saturation with an accumulation of risk factors can considerably increase the probability of needing an ICU bed, thus, leaving the healthcare team to make the appropriate decisions.

\subsubsection{Scenario Five: Investigating the Vaccines used in the Brazilian Immunization Program}

Vaccination in Brazil against the new coronavirus began on January 17, 2021, applying an initial batch of an attenuated virus-based vaccine called CoronaVac, developed within the collaboration of the Chinese company Sinovac Biotech and Butantan Institute  \cite{fonseca21covidvacc}. Six days later, another immunizer has been added to the vaccination program, the AZD1222, which is based on adenovirus and developed by Oxford University and AstraZeneca. These two immunizers have been widely applied in the Brazilian campaign. Finally, in late April, a third immunizer has been added to the program; the mRNA-based COVID-19 vaccine named Comirnaty developed by Pfizer-BioNTech.

As in many other cases worldwide, the Brazilian vaccination program has faced severe problems like shortages of doses and supplies for vaccines \cite{hallal21overcoming}. In this way, the program's accomplishment takes place according to the availability of each mentioned vaccine, without a defined strategy for targeting the different vaccines types. The program holds primarily three different vaccines developed by three different technologies, and in an ideal world, these immunizers would be destined at key groups aiming at their optimal use. In this scenario, we have used our tool to understand the behavior of these diverse vaccines throughout the Brazilian vaccination program.

\begin{figure}
	\centering
	\includegraphics[width=13 cm]{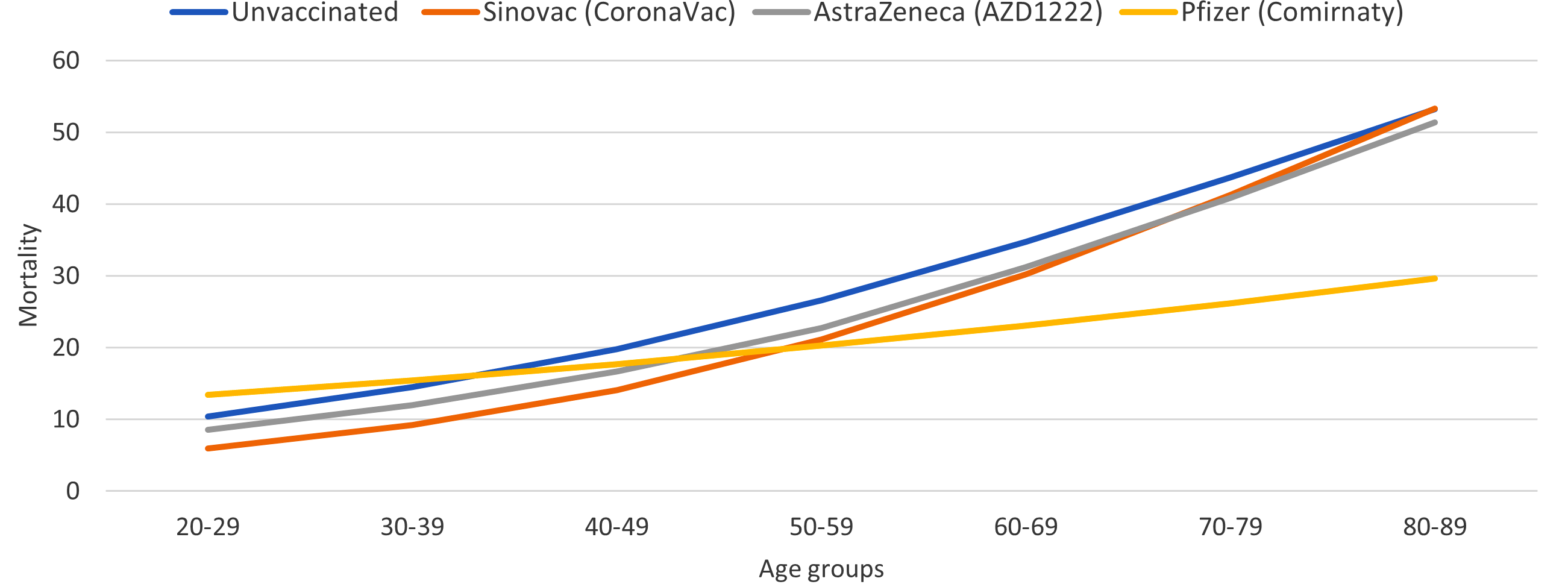}
	\caption{Estimates of mortality rates by age groups according to the logistic regression analysis. Although the performance of vaccines seems similar in younger people, we cannot make any statements about it due to the lack of data in the phases before hospitalization. However, the superiority of the Comirnaty vaccine in older groups seems to be explicit.}
	\label{fig:vaccines} 
\end{figure}

The data was obtained from the DATASUS (Department of Informatics of the Unified Health System) \footnote{https://datasus.saude.gov.br/}. This Brazilian public database collects and makes available information from the public health system. The data set gathers information on Severe Acute Respiratory Syndrome (SARS) admissions, including data from SARS-CoV-2, from January 1 to July 12, 2021. All data is appropriately anonymized.

Firstly, we have split the data set between unvaccinated and vaccinated for each of the used vaccines with at least one shot. We then open the subsets in our tool to generate the regression models by looking at the label "death ." An interesting initial observation is when we look for correlations of attributes to death cases, the subsets of CoronaVac and AZD1222 show significant correlations with age (respectively 0.21 and 0.26), unlike the Comirnaty's subset, which has an irrelevant correlation (0.03). This gives us evidence that Pfizer's vaccine is less sensitive to age differences keeping its efficacy more stable across age groups.

Figure \ref{fig:vaccines} shows how logistic regression learned the characteristic of each vaccine with death as the target label and age range as the selected attribute. Unfortunately, since the data deal with hospitalization cases, we cannot investigate the performance of vaccines before this moment. That said, the figure suggests that for young people, vaccines tend to be equally effective for hospitalization cases. However, for elderly patients, there is a clear advantage to those who have received doses of Pfizer's vaccine.

\begin{figure}
	\centering
	\includegraphics[width=13 cm]{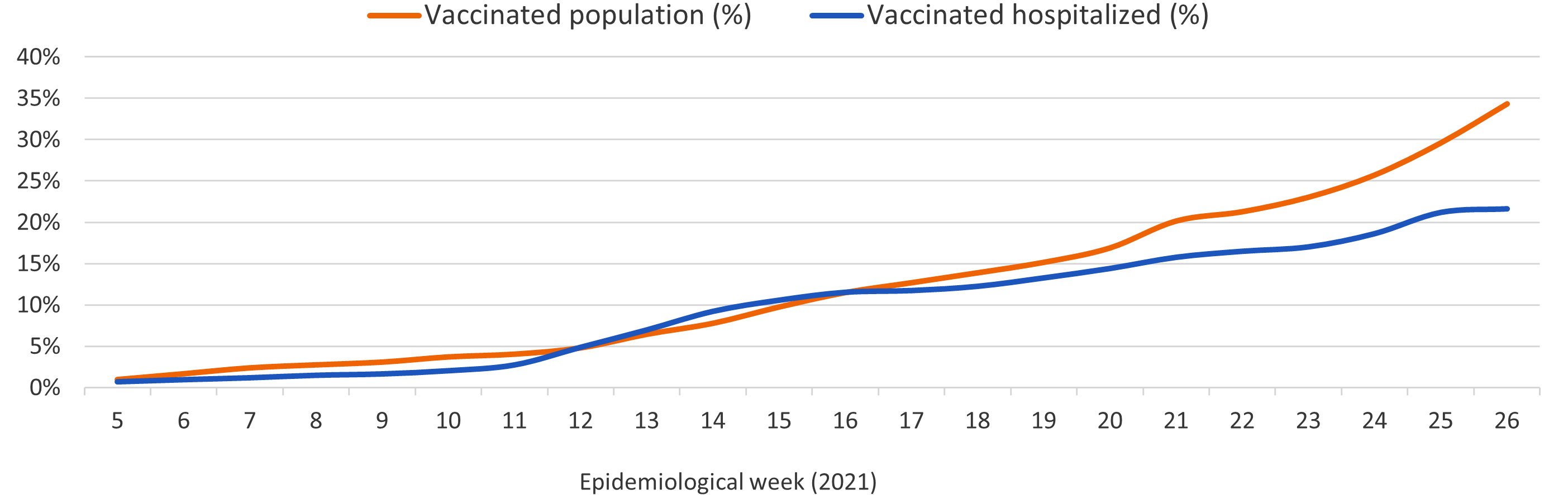}
	\caption{Comparison of the population vaccination rate and the rate of vaccinated ones among hospitalized patients. Around the seventeenth week of the year, those vaccinated started to represent a smaller proportion among hospitalized patients.}
	\label{fig:campaign} 
\end{figure}

Another pertinent observation is that for hospitalizations of older people (manly over 70), both those vaccinated with CoronaVac and AZD1222 do not seem to have any advantage over non-vaccinated ones. And these are precisely the vaccines mainly applied to these groups, which may represent an initial failure of the program. To support this statement, we have searched for data about the progress of vaccination in Brazil in the repository Our World In Data \footnote{https://ourworldindata.org/}. As shown in Figure \ref{fig:campaign}, the proportion of those vaccinated becomes significantly lower from the seventeenth week onwards, approximately when vaccination reached the 65-year-olds.

\section{Conclusions}

In this paper, we have presented a novel visual analytics approach and its simple and portable prototype for generating, evaluating, and applying regression models. The goal is to allow final users to create models with their data sets and make use of them in a unified tool. The approach works by initially exposing a panorama of correlations between attributes and a target event (or outcomes), thus allowing users to choose good attributes for creating regression models. To formulate a definitive and reliable model, UCReg  combines visual methods to evaluate the generated models; hence, the users can gain insights about the data and proceed with the production of the finalized model for later use.

Among the contributions of our work, we highlight the development of complete visual analytics approach to the problem of building, evaluating and  a freely available web-based prototype for the transparent exploration of regression models. Also, the development of LoRRViz,  an adapted RadViz as a visual evaluation approach for multinomial LR models. It comes as an alternative to the classical ROC curves, or even to confusion matrices, once it generates a visual overview of how the model is classifying items.

At present, some limitations of our work still stand. The approach deals only with numerical attributes as independent attributes. Given the current high availability of categorical data, it would be interesting to make them available so that the user can perform regressions and consult the impact of these attributes on predictions. Another limitation relates to the target attribute (dependent attribute); it must encode meaningful states of the data set. Poorly labeled target attributes may bring inaccuracy to the generated models. Finally, the approach scalability potential (once attributes are plotted in a point-based visualization); however, our prototype is implemented in Javascript with the D3js library, which does not imply the best performance option. Yet, in our tests, we were able to interactively generate models with hundreds of attributes with items in the ten thousand mark.

\bibliographystyle{unsrtnat}
\bibliography{references}  






\end{document}